\documentclass[12pt]{article}

\usepackage{newtxtext,newtxmath}

\usepackage{subfigure}
\usepackage{booktabs}
\usepackage{amsmath}
\usepackage{mathtools}

\usepackage{bm}
\usepackage{eqnarray}    
\usepackage{graphicx}    

\usepackage[a4paper, total={6.5in, 9in}]{geometry}
\usepackage[colorlinks=true,linkcolor=black,anchorcolor=black,citecolor=blue,filecolor=black,menucolor=black,runcolor=black,urlcolor=black]{hyperref}

\usepackage[square, numbers]{natbib}

\usepackage{tabularx}
\newcolumntype{L}{>{\centering\arraybackslash}m{3cm}}

\usepackage[usestackEOL]{stackengine}

\usepackage{multirow}

\usepackage[width=.875\textwidth]{caption}
\usepackage{adjustbox}

\usepackage{mathpazo}                      
\usepackage[small, euler-digits]{eulervm}  

\usepackage[usestackEOL]{stackengine}

\usepackage{authblk}

\renewcommand{\small}{\fontsize{12}{14.4}\selectfont}

\renewcommand{\large}{\fontsize{14}{16.8}\selectfont}
\renewcommand{\Large}{\fontsize{18}{21.4}\selectfont}

\usepackage{titlesec}

\titleformat*{\section}{\Large\bfseries}
\titleformat*{\subsection}{\large\bfseries}
\titleformat*{\subsubsection}{\large\bfseries}

\begin{document}

\title{Unsupervised Optimisation of GNNs for Node Clustering}
\author{William Leeney and Ryan McConville}
\affil{School of Engineering Mathematics and Technology, University of Bristol}

\maketitle

\abstract{Graph Neural Networks (GNNs) can be trained to detect communities within a graph by
learning from the duality of feature and connectivity information. Currently, the common approach for optimisation of GNNs is to use comparisons to ground-truth for hyperparameter tuning and model selection. In this work, we show that nodes can be clustered into communities with GNNs by solely optimising for modularity, without any comparison to ground-truth. Although modularity is a graph partitioning quality metric, we show that this can be used to optimise GNNs that also encode features without a drop in performance. We take it a step further and also study whether the unsupervised metric performance can predict ground-truth performance. To investigate why modularity can be used to optimise GNNs, we design synthetic experiments that show the limitations of this approach. The synthetic graphs are created to highlight current capabilities in distinct, random and zero information space partitions in attributed graphs. We conclude that modularity can be used for hyperparameter optimisation and model selection on real-world datasets as well as being a suitable proxy for predicting ground-truth performance, however, GNNs fail to balance the information duality when the spaces contain conflicting signals.}

\newpage
\section{Introduction}

GNNs can be trained to learn representations which are useful for the detection of communities of nodes. The task of attributed graph clustering, where nodes have associated features, is to encode pertinent information from both the adjacency and feature spaces to create partitions of similar nodes. GNNs combine these two sources of information by propagating and aggregating node feature embeddings along the network's connectivity \citep{scarselli2008graph, kipf2016semi}. GNNs can be trained to learn representations that are useful for node clustering without using labels in the loss function. However, currently the typical approach when training unsupervised GNNs is to use labelled information for model selection and the early stopping criteria, which is a form of supervision \citep{xie2022self,JMLR_DIG,khoshraftar2022survey,waikhom2023survey,rahman2021comprehensive}. Relying on labels to perform unsupervised learning means that we cannot train models in scenarios with no ground-truth available. 

Many real-world systems are best represented with a graph-structure and unsupervised learning has numerous practical applications. In the cybersecurity domain, where the computer network is a graph, unsupervised learning can be used for intrusion detection where unauthorised parties are unknown \citep{yousefi2017autoencoder}. If unsupervised algorithms perform well on the ground-truth of a task without being trained on the ground-truth, they will also perform well in scenarios where the ground-truth is not available. Similarly, this is also applicable to misinformation detection \citep{nielsen2022mumin} or genomic feature discovery \citep{cabreros2016detecting} where the ground-truth is not known as it is hard to collect or annotate. Additionally, in academic citation networks, predicting the influence or impact of research papers can be challenging without labeled data on future citations. Unsupervised methods can be used to extract meaningful features from citation structures \citep{chen2010community}, helping researchers assess the potential significance of a paper, its contribution to a field, and its likelihood of future citations.

If the ground-truth is not available for an application, it is useful to know if unsupervised metrics of clustering quality can be used to select models that would perform well on a given task. Therefore, in this work we investigate whether unsupervised graph partitioning metrics such as modularity and conductance, can be used in place of the ground-truth for model selection. Further, we investigate if the ground-truth performance can be estimated using these metrics as a proxy. The problem of unsupervised model selection has been investigated by \citet{duan2019unsupervised} but only in the context of variational autoencoders that learn disentangled representations. Here we investigate a range of GNNs that are not specifically designed to disentangle the latent space. Our investigation addresses whether these GNNs can be optimised without any comparison to ground-truth by comparing multiple instances of the same architectures across different random seeds and quantifying how correlated unsupervised metrics' performance is with measures of ground-truth.  

When training models, it is important to select the right hyperparameters for a task, as previous work has shown that the HPO significantly affects performance \citep{yang2020hyperparameter,bischl2021hyperparameter,ugle2023leeney,palowitch2022graphworld,dwivedi2020benchmarking} and the best hyperparameters for each method depend on the dataset \citep{chen2022bag}. Choosing the best hyperparameters for a dataset is an arduous task which is typically performed with comparisons to a validation set of ground-truth data. The hyperparameters that are used by default are those specified by the original authors and it is likely that these were found with comparisons to the ground-truth. Therefore, in this work, we investigate if unsupervised metrics can be used to evaluate the hyperparameters that lead to good clustering performance.

Another important model-agnostic factor in achieving good model performance is the size of the dataset. Unfortunately, there is often a low labelling rate with datasets that mimic realistic scenarios as labels can be expensive to gather \citep{khatua2023igb,hu2021ogb}.  There is previous research that studies how the structure of data contributes to the  generalization bounds from using limited training data \citep{singh2008unlabeled, epstein2019generalization}. \citet{epstein2019generalization} introduce a theoretical bound and compare with an empirical observed test loss for unsupervised learning with autoencoders, whereas our work is solely an empirical study when using unsupervised optimisation for graph representation learning. \citet{singh2008unlabeled} theoretically and \citet{le2018supervised} empirically study the lower bounds of situations where unsupervised learning contributes to the improvement of supervised learning, whereas we compare the performance from a reduced training set with the performance of a supervised optimisation. It is useful to study the lower limit of data needed to optimise GNNs with unsupervised metrics, to understand how a model may perform and so that it can be applied on data outside of the training dataset.

Our experiments use existing metrics of clustering partition quality in the adjacency matrix to see if this correlates with the clusters that appear in GNNs learnt representation space. We investigate this because we are studying attributed graph clustering, and we can have feature space information as well as the adjacency connectivity. We assume that the adjacency space clustering quality reflects good quality clusters in the feature space, as the ground-truth will reflect both information spaces. Therefore, it is pertinent to examine why unsupervised GNN models can be trained with unsupervised graph metrics alone. We design synthetic data experiments where the adjacency space reflects the feature space as well as when the spaces contain conflicting information so that we can examine the performance limitations and explain why unsupervised metrics can be used as a proxy for ground-truth.

In this work, we use GNNs that can learn representations without using labels in the loss function and optimise them with the unsupervised metrics of clustering, modularity and conductance. We then compare performance to metrics of ground-truth to evaluate the validity of our approach. A wide selection of algorithms are tested over many datasets. The $W$ randomness coefficient is used as the metric of trust in the findings by quantifying the consistency of rankings over random seeds in experiments \citep{ugle2023leeney}. In summary, this paper answers the following research questions for GNN clustering algorithms:

\begin{itemize}
    \item RQ1: If GNNs are optimised without any comparison to ground-truth, can the performance on unsupervised metrics serve as an approximation for evaluating against ground-truth labels? 
    \item RQ2: Can GNNs be trained without ground-truth comparisons and a significant drop in performance, compared with training GNNs on ground-truth comparisons?
    \item RQ3: Can unsupervised metrics be used for the optimisation of hyperparameters for GNNs and will performance still be correlated with the ground-truth performance?
    \item RQ4: What is the lower limit of data needed for unsupervised optimisation of GNNs in these settings?
    \item RQ5: How does a synthetic generation of clustering partitions in both feature and adjacency spaces help us understand unsupervised GNN model selection?
\end{itemize}

\section{Related Work}

Several frameworks for the evaluation of supervised GNNs performing node classification, link prediction and graph classification exist \citep{dwivedi2020benchmarking,morris2020tudataset,errica2019fair,palowitch2022graphworld}. We are interested in unsupervised community detection which is more challenging to train and evaluate due to the lack of any supervision signal. There are plenty of reviews that describe the different approaches to community detection \citep{liu2020deep,jin2021survey,saxena2017review} but they do not empirically compare the performance or discuss how GNNs can be selected for without a validation set of ground-truth. \citet{cappelletti2023grape} present a library that is efficient for large scale graph investigations and they compare methods for random-walk based graph representation learning. However, they use validation techniques for model selection that include a supervised comparison to evaluate models, whereas in our work we use none. The study we carry out is specific to learning with GNNs on attributed graphs and does not compare to non-neural clustering methods such as k-means \citep{sinaga2020unsupervised} or Louvain \citep{blondel2008fast} as GNNs have been shown to outperform more traditional methods \citep{khanfor2020graph,sobolevsky2022graph,rahman2021comprehensive}.

\citet{dinh2015network} investigates the theoretical performance bounds of using modularity to optimise algorithms in finding community structure, whereas here we look at how modularity can be used to find the structure defined by ground-truth. Similarly, both \citet{akbiyik2023graphtester} and \citet{d2021new} explore the best theoretical performance that can be achieved with a GNN on a given dataset, whereas this work investigates the limit of current methods when evaluating with unsupervised metrics. We quantify to what extent communities can be detected with GNNs algorithms alone rather than try to reduce uncertainty in clustering results as done by \citet{christakis2023reducing} and \citet{esfandiari2023replicable}.

The investigations performed by \citet{locatello2020sober} finds that there is no evidence that models can be trained in a unsupervised manner for disentangled representations as the random seeds and hyperparameters have a significant effect. However, this is unclear as \citet{duan2019unsupervised} show that it is possible to perform unsupervised model selection for disentangling models by performing pairwise comparisons between trained model representations across different seeds. The solution only works for this class of model because the proposed unsupervised disentanglement ranking metric quantifies how similar two embedding spaces are by measuring how disentangled the embedding space is. This is possible because the theoretical work of \citet{rolinek2019variational} says for a given non-adversarial dataset, a disentangling VAE will likely keep converging to the same disentangled representation. This is in contrast to non-disentangling neural networks which when trained with the same hyperparameters and architectures converge to different embedding space representations for the same data whilst still maintaining similar performance \citep{li2015convergent,wang2018towards,morcos2018insights}. Therefore, their work is not appropriate for non-disentangling neural networks, whereas we investigate if it is possible to train unsupervised models despite the effect of hyperparameters and random seeds, but none of the models studied in our work are required to learn disentangled representations.

We do not consider the ability of transfer learning \citep{sener2016learning} for transferable representations, as in our research learning is conducted directly on the target domain as to do otherwise assumes prior knowledge. Using other datasets means that other sources of ground-truth are needed, whereas we want to train models in situations with no ground-truth comparisons. Similarly, \citet{zhao2021automatic} used historical dataset performance for unsupervised model selection for outlier detection however, our research doesn't use any ground-truth even from previous datasets. \citet{ma2023need} does not use any ground-truth as they use internal evaluation strategies for model selection that are only based on the dataset being trained on, however, this is for outlier detection not GNN clustering. \citet{erhan2010does} studies how unsupervised pre-training helps supervised models, which illustrates that unsupervised learning is useful, but we study how well unsupervised learning performs without any additional training on top. 

In this work, we investigate the lower limit of data needed for unsupervised optimisation of GNNs. This is similar to \citet{ashtiani2015representation} who provide a formal model for assessing if a relatively small random sample of a dataset will lead to convergence of k-means clustering, but ours is an empirical analysis of GNN clustering. This is different to \citet{zhu2021shift} who try to alleviate potential data distribution bias in the training data by designing a GNN that more effective beyond the data it is trained on. We do not investigate the practical limitations of different GNN architectures as done by \citet{alon2020bottleneck} or ways of speeding up training process like \citet{wang2021empirical} rather, we look removing a labeled comparison for model validation. We do investigate what types of structures our approach can be used for which is different to \citet{magner2019fundamental} for they show the conditions under which supervised GCNs are capable of distinguishing between sufficiently well separated graphons, whereas the synthetic experiments in this work investigates different unsupervised GNNs ability to separate informative structure in the connectivity and feature spaces. There does exist previous work that generates synthetic graphs to evaluate GNNs \citep{palowitch2022graphworld}, but this focuses on assessing the landscape of current graph datasets used for benchmarking with respect to the potential datasets and what they show. \citet{tsitsulin2022synthetic} looks at how sensitive clustering GNNs are to parameters that are used to generate random graphs whereas our work investigates if unsupervised model selection can balance clustering partitions in both connectivity and feature spaces.

\section{Methodology}

This section details the clustering problem and the experiments that will answer the research questions outlined previously. The framework procedure is to use conductance ($\mathcal{C}$) and modularity ($\mathcal{M}$) as the only metrics for model evaluation to keep the process completely unsupervised. The supervised metrics that are predicted are Marco-F1 ($F1$) and $NMI$. All experiments are repeated over ten random seeds and to quantify the sensitivity of the results to randomness we use the $W_w$ (denoted hereafter as $W$) randomness coefficient \citep{leeney2023uncertainty} which has been shown to be the best regardless of investigation size. 

The focus of this investigation is to understand if it is possible to train unsupervised GNNs with no information on the ground-truth of a task and show that an unsupervised measure of performance is correlated with metrics of ground-truth (RQ1) whilst still performing well on said ground-truth (RQ2). This is useful for applications where you want to train a model but don't have ground-truth to guide model selection. To show that this is possible, the models are optimised with both modularity and conductance independently and then we compute the correlation with $NMI$ and $F1$. A strong correlation will indicate that the labelled performance can be predicted from the unsupervised performance (RQ1). The correlation is used to estimate the performance of each algorithm on the ground-truth, so that practitioners can train unsupervised models and use them with appropriate reliability for a task before labels have been gathered.

To answer RQ1, can the performance a GNN trained without supervision on unsupervised metrics serve as an approximation for evaluating against ground-truth, each GNN is trained with the default hyperparameters and all available data. To perform model selection of the best epoch to evaluate the model on, each is selected based on best performance on modularity and conductance. To quantify if performance on the unsupervised metrics is correlated with the supervised metrics, the optimised model is then compared with the ground-truth and the $R^2$ coefficient is used to compute correlation. To assess the performance difference between an unsupervised model selection and a labelled one (RQ2), we also report the mean absolute difference ($MAE$) in performance that could've been achieved if we had used the labelled information for model selection. 

The caveat with the first investigation is that the default hyperparameters may have been found by the original authors using a comparison to the ground-truth labels for model selection. Choosing these is not trivial and requires a thorough analysis, which is typically performed with comparisons to a validation set of labels. For this reason, we study if unsupervised metrics can be used to optimise the hyperparameters, with the additional challenge of no knowledge of good hyperparameters (RQ3). The modification in this experiment is to carry out a hyperparameter optimisation for each dataset and algorithm combination using modularity and conductance. This is a more realistic scenario and is important as it has been shown that the HPO significantly affects the ranking of algorithms at this task \citep{ugle2023leeney,palowitch2022graphworld} and that different hyperparameters are important for different datasets \citep{bergstra2012random,yang2020hyperparameter}. 

Additionally, to test real world applicability it is pertinent to investigate the lower limit of data needed for an unsupervised optimisation pipeline (RQ4). To test this, we investigate if training on a subset of the original dataset still correlates with the performance on the full dataset available. To do this, the default hyperparameters are used to train the models on 33\% and 66\% of the data. These models are then evaluated with the labels on the full dataset. We then quantify the correlation between the reduced dataset performance with unsupervised metrics and the labelled performance on all of the data. This is useful as often models will be used in scenarios that they have not been explicitly trained for, as well as not having a supervision signal of ``goodness''.

We are examine whether GNN exhibit bias towards either information space, adjacency or feature. We want to investigate GNN capabilities when there is a distinct clustering signal in both information spaces compared to when there is no cluster separation. Therefore, we create synthetic data with three different levels of cluster-able information. The three different partitions can be created in both feature and adjacency spaces and are ``distinct", ``random" and ``null". So, using a synthetic generation of the clustering partitions in both feature and connectivity spaces, we study why training GNNs to cluster with unsupervised metrics work (RQ5). 

The ``distinct" scenario for connectivity is that all nodes within the cluster are connected and there are no intra-cluster connections. For the ``distinct" feature space, each cluster has half of the features assigned as one and the rest zero, with the reverse true for the opposite cluster. The ``distinct" scenario is the strongest clustering signal that can be generated in each space. The ``random" scenario for both spaces randomly allocates edges and features, which gives no signal that indicates a difference between clusters. The ``null" scenario for the graph is that all potential edges are present and for the feature space, every node has all features present. The ``null" partition indicates that there is only one cluster, which is the opposite of the ``distinct" scenario. Each possible combination of these splits between clusters is tested to see if the unsupervised optimisation balances the duality of the attributes and connections of the nodes. This is done to see if the GNNs are able to detect that there is no information to be gained and can is therefore aware that the clustering is only present on the other spaces' information. The datasets generated are 1000 nodes with 500 features with two clusters both sized 500 nodes.

\subsection{Problem Definition}\label{section: problem_def}

The problem definition of community detection on attributed graphs is defined as follows. The graph, where $N$ is the number of nodes in the graph, is represented as $G = (\textbf{A}, \textbf{X})$, with the relational information of nodes modelled by the adjacency matrix $\textbf{A} \in \mathbb{R}^{N \times N}$. Given a set of nodes $V$ and a set of edges $E$, let $e_{i, j} = (v_i, v_j) \in E$ denote the edge that points from $v_j$ to $v_i$. The graph is considered undirected and unweighted so, the adjacency matrix $A_{i, j} = 1$ if $e_{i,j} \in E$  and $A_{i, j} = 0$ if $e_{i,j} \notin E$. Also given is a set of node features $\textbf{X} \in \mathbb{R}^{N \times d} $, where $d$ represents the number of different node attributes (or feature dimensions). The objective is to partition the graph $G$ into $k$ clusters such that nodes in each partition, or cluster, generally have similar structure and feature values. The only information given to the algorithms is the number of clusters $k$ to partition the graph into. Hard clustering is assumed, where each community detection algorithm must assign  each node a single community to which it belongs, such that $\textbf{P} \in \mathbb{R}^{N}$. There exists ground-truth associated with each dataset, such is that each node has an associated cluster $\textbf{L} \in \mathbb{R}^{N}$, this is not used at all in the training optimisation. 

\subsection{Datasets}

To fairly compare different GNN architectures, a selection of graph datasets are used to represent some potential applications. Each dataset can be summarised by commonly used graph statistics: the average clustering coefficient \citep{watts1998collective} and closeness centrality \citep{wasserman1994social}. The former is the proportion of all the connections that exist in a nodes neighbourhood compared to a fully connected neighbourhood, averaged across all nodes. The latter is the reciprocal of the mean shortest path distance from all other nodes in the graph. All datasets are publicly available\footnote{\url{https://github.com/yueliu1999/Awesome-Deep-Graph-Clustering}}, and have been used previously in GNN research \citep{deep_graph_clustering_survey}. These datasets are detailed in Table \ref{tab:stats}, the following is a brief summary: Cora \citep{mccallum2000automating}, CiteSeer \citep{giles1998citeseer}, DBLP \citep{tang2008arnetminer} are graphs of academic publications from various sources with the features coming from words in publications and connectivity from citations. Texas, Wisc and Cornell are extracted from web pages from computer science departments of various universities \citep{craven1998learning}. Computers and Photo \citep{shchur2018pitfalls} are co-purchase graphs extracted from Amazon, where nodes represent products and edges the co-purchases, and features are bag-of-words vectors extracted from product reviews.

\setlength\tabcolsep{3.5pt}
\begin{table}[ht]
\centering
\small
\captionsetup{font=small} 
\begin{tabular}{@{}c|cccccc@{}}
\toprule
\textbf{\Longunderstack{\\Datasets}}  & \Longunderstack{\\Nodes} & \Longunderstack{\\Edges} & \Longunderstack{\\Features} & \Longunderstack{\\Classes} & \Longunderstack{Average\\Clustering\\ Coefficient} & \Longunderstack{Mean\\Closeness\\ Centrality} \\
\midrule
\midrule
CiteSeer  \citep{giles1998citeseer} & 3327 & 9104 & 3703 & 6 & 0.141 & 0.045 \\
Cora \citep{mccallum2000automating} & 2708 & 10556 & 1433 & 7 & 0.241 & 0.137 \\
DBLP \citep{tang2008arnetminer} & 4057 & 7056 & 334 & 4 & 0.177 & 0.026 \\
Texas \citep{craven1998learning} & 183 & 325 & 1703 & 5 & 0.198 & 0.344 \\
Wisc \citep{craven1998learning}  & 251 & 515 & 1703 & 5 & 0.208 & 0.32 \\
Cornell \citep{craven1998learning} & 183 & 298 & 1703 & 5 & 0.167 & 0.326 \\
Computers \citep{shchur2018pitfalls} & 13752 & 491722 & 767 & 10 & 0.344 & 0.286 \\
Photo \citep{shchur2018pitfalls}  & 7650 & 238162 & 745 & 8 & 0.404 & 0.242 \\
\bottomrule
\end{tabular}

\caption{The datasets and associated statistics.}\label{tab:stats}
\end{table}
\noindent

\subsection{Models}

In addition to GNNs designed specifically for community detection GNNs, we also consider those that can learn an unsupervised representation of data, as there is existing research that succesfully applies vector-based clustering algorithms to the representations \citep{fard2020deep,wang2021decoupling}. We open-source the code necessary to train all of the GNN models considered\footnote{\url{https://github.com/willleeney/ugle}}. Deep Attentional Embedded Graph Clustering (DAEGC) uses a k-means target to self-supervise the clustering module to iteratively refine the clustering of node embeddings \citep{wang2019attributed}. Deep Modularity Networks (DMON) uses GCNs to maximises a modularity based clustering objective to optimise cluster assignments by a spectral relaxation of the problem \citep{tsitsulin2020graph}. Deep Graph Infomax (DGI) maximises mutual information between patch representations of sub-graphs and the corresponding high-level summaries \citep{velickovic2019deep}. GRAph Contrastive rEpresentation learning (GRACE) generates a corrupted view of the graph by removing edges and learns node representations by maximising agreement across two views \citep{zhu2020deep}. Bootstrapped Graph Latents (BGRL) \citep{thakoor2021bootstrapgraph} uses a self-supervised bootstrap procedure by maintaining two graph encoders; the online one learns to predict the representations of the target encoder, which in itself is updated by an exponential moving average of the online encoder. Towards Unsupervised Deep Graph Structure Learning (SUBLIME) \citep{liu2022towards} uses two views encoder with the bootstrapping principle applied over the feature space as well as a contrastive scheme between the nearest neighbours. Variational Graph AutoEncoder Reconstruction (VGAER) \citep{qiu2022VGAER} reconstructs a modularity distribution using a cross entropy based decoder from the encoding of a VGAE \citep{kipf2016variational}. 

\setlength\tabcolsep{7pt}
\begin{table}[h!b]
\centering
\small
\captionsetup{font=small} 
\begin{tabular}{@{}c|c@{}}
\toprule
Resource & Associated Allocation  \\
\midrule
\midrule
Optimiser & Adam \\
Max Epochs & $5000$ \\
Max Hyperparameter Trials & $250$ \\
Seeds & $\{42, 24, 976, 12345, 98765, 7, 856, 90, 672, 785\}$ \\
Learning Rate & $\{0.05, 0.01, 0.005, 0.001, 0.0005, 0.0001\}$ \\
Weight Decay & $\{0.05, 0.005, 0.0005, 0.0\}$ \\
Patience & $\{25, 100, 500, 1000\}$ \\

\bottomrule
\end{tabular}
\caption{Resources allocated to the hyperparameter investigation, those detailed are shared across all investigations. Algorithms that are designed to benefit from a small number of HPs should perform better as they can search more of the space within the given budget. All models are trained with 1x 2080 Ti GPU and a 16core Xeon CPU.}\label{tab: resources}
\end{table}
\noindent

\subsection{Hyperparameter Optimisation Procedure}

There are a variety of Bayesian methods that can be used for hyperparameter selection but given that the Tree Parzen-Estimator (TPE) \citep{bergstra2013making, ozaki2020multiobjective} can retain the conditionality of variables \citep{yang2020hyperparameter} and has been shown to be a efficient estimator given limited resources \citep{yuan2021systematic}, we will use the TPE for the framework evaluation environment. As we are investigating multiple unsupervised metrics of performance, the multi-objective version of the TPE \citep{ozaki2020multiobjective} is used. Where relevant, the hyperparameters are optimised on each seed used.

\section{Results}


We quantify the correlation between the unsupervised and supervised metrics using the $R^2$ coefficient of determination with both a linear ($l-R^2$) and quadratic model ($q-R^2$) to measure what percentage of variation is explained by the models. We record the absolute performance of all experiments as well as the $MAE$ which is the mean absolute difference from using the supervised metrics in the optimisation process. The ranking randomness sensitivity present in the investigation is quantified with $W$ randomness coefficient.

\subsection{Unsupervised Model Selection and Hyperparameter Optimisation (RQ1-3)}

\begin{figure*}[hp!]
    \centering
    \captionsetup{font=small} 
    \begin{minipage}[b]{\textwidth}
        \centering
        \includegraphics[width=\textwidth,height=0.37\textheight]{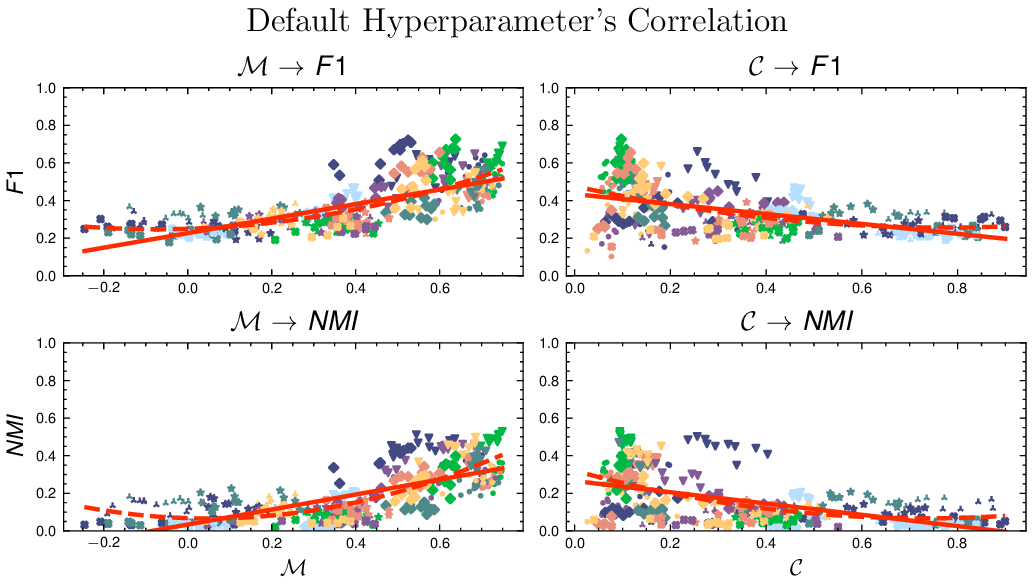}
    \end{minipage}
    \vspace{0.8cm}
    \begin{minipage}[b]{\textwidth}
        \centering
        \captionsetup{font=small} 
        \includegraphics[width=\textwidth,height=0.37\textheight]{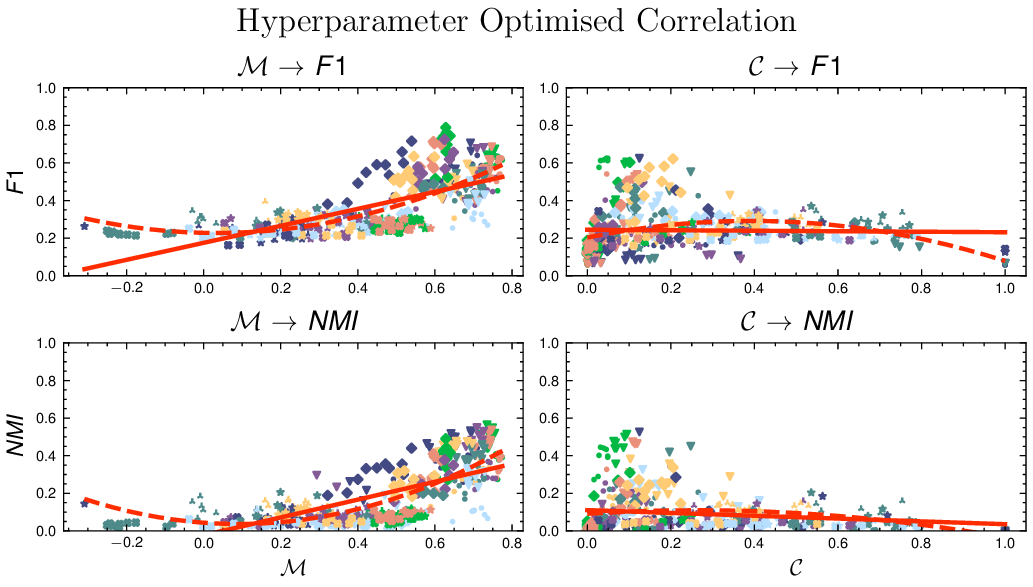}
        \includegraphics[height=0.07\textheight]{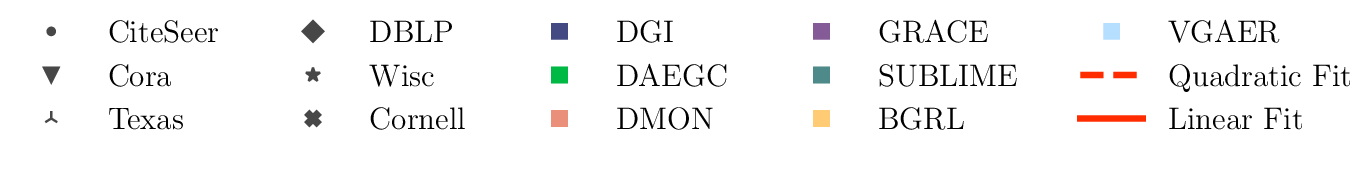}
        \caption{A visualisation of the correlations between the performance of GNN models using default hyperparameters when using modularity and conductance for model selection against the ground-truth metrics. Both linear and quadratic curves are fitted to each pair of metrics. Also presented here are the results of an unsupervised hyperparameter optimisation using modularity and conductance as the performance metrics. The optimised models are compared with the ground-truth performance on $F1$ and $NMI$.}
        \label{fig:hpo_def_corr}
    \end{minipage}
\end{figure*}

\begin{figure*}[hbt!p]
    \centering
    \captionsetup{font=small} 
    \begin{minipage}[b]{\textwidth}
        \includegraphics[width=\textwidth,height=0.38\textheight]{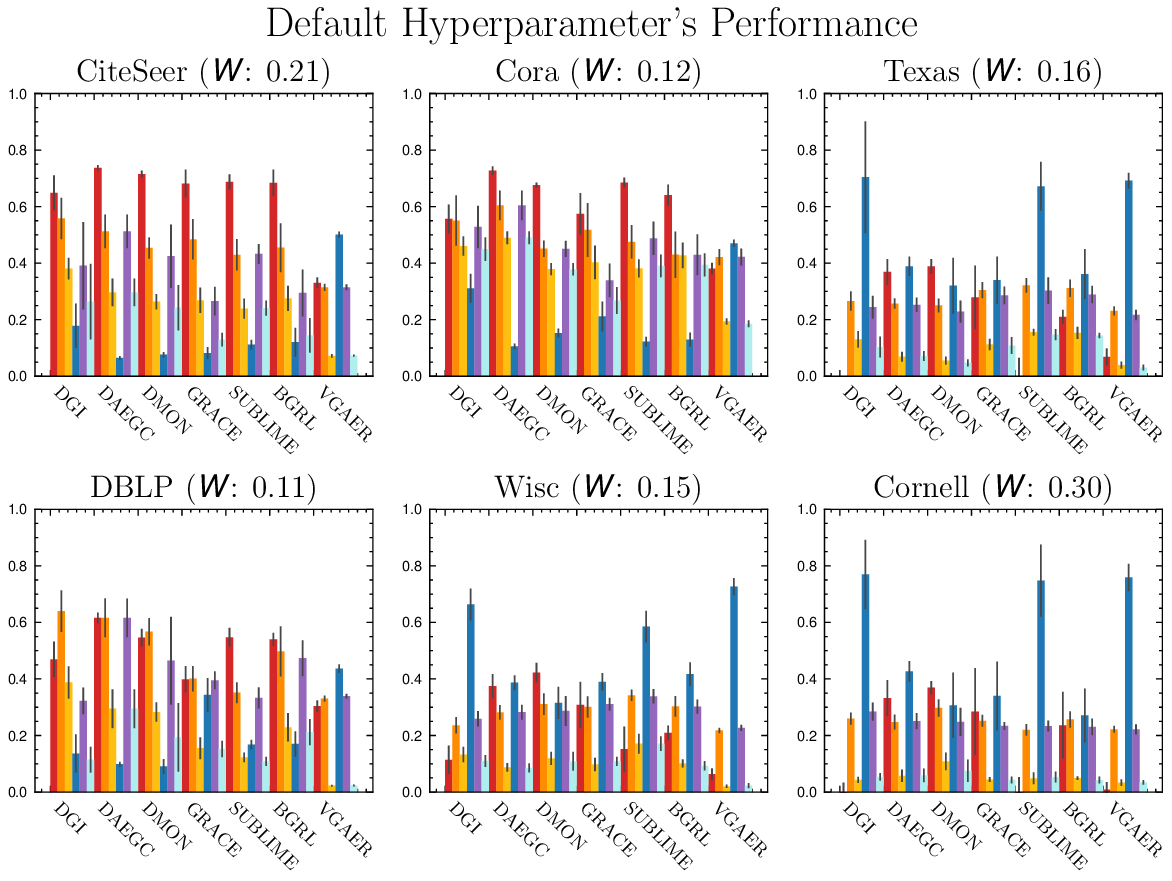}
  \end{minipage}
  \vspace{0.8cm}
  \begin{minipage}[b]{\textwidth}
    \centering
    \includegraphics[width=\textwidth,height=0.38\textheight]{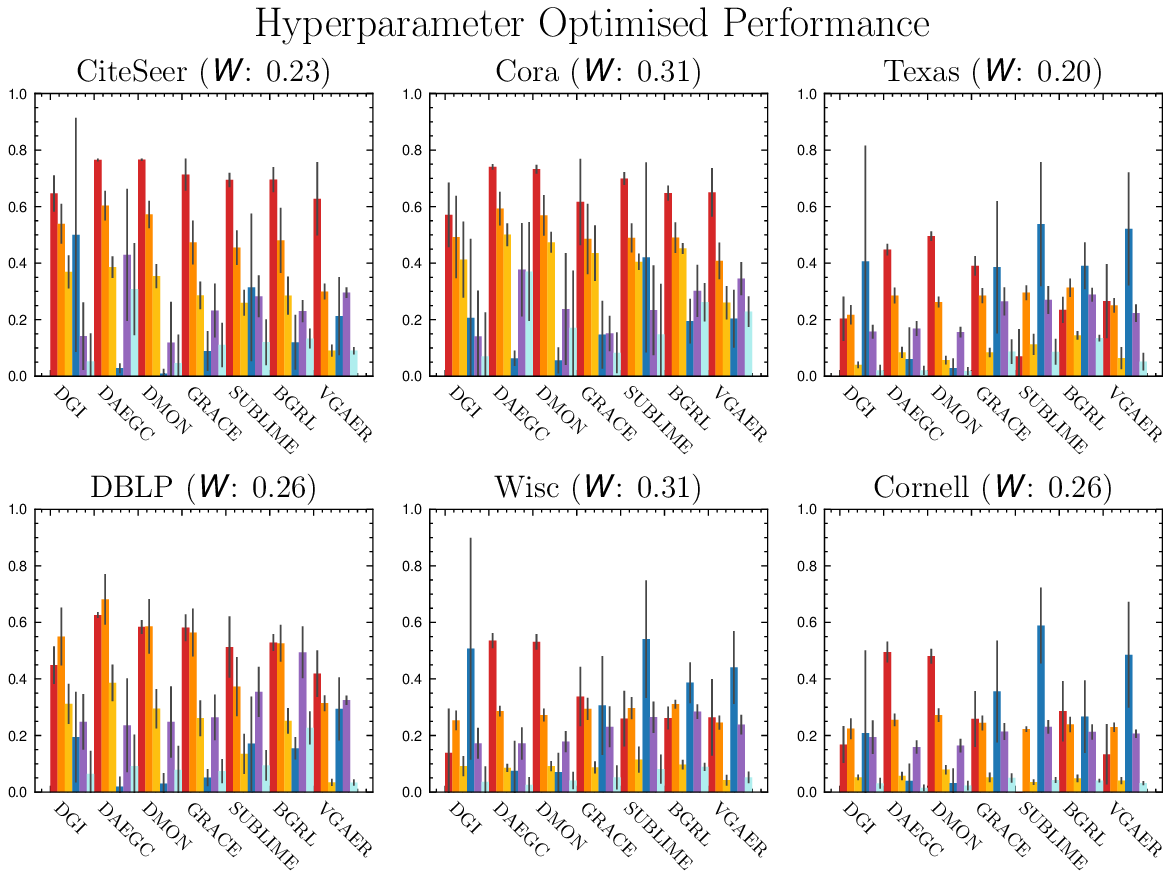}
    \includegraphics[height=0.07\textheight]{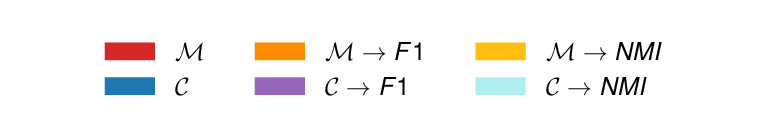}
    \caption{The absolute performance of each algorithm on the ground-truth metrics F1 and NMI when using the default hyperparameters and performing model selection based on modularity and conductance. We also visualise the performance of each algorithm on each metric when optimising hyperparameters of the GNN models on modularity and conductance. The deviation due to random seed is given by the error bars. }
    \label{fig:def_hpo_abs}
  \end{minipage}
\end{figure*}

We first discuss the use of unsupervised metrics for model selection and HPO, where Table \ref{tab:def_hpo_overall_res} summarises all of the results over every dataset and algorithm in these experiments as well as details the framework comparison rank ($FCR$). In Figure \ref{fig:hpo_def_corr} the correlations are visualised for the default hyperparameters and the unsupervised HPO whereas Figure \ref{fig:def_hpo_abs} shows the absolute performance comparison.

\renewcommand{\arraystretch}{1.2}
\setlength{\tabcolsep}{15pt}
\begin{table}[h!tb]
    \centering
    \small
    \captionsetup{font=small} 
    \begin{tabular}{c|cccc}
        \toprule
        \multirow{2}{*}{\Longunderstack{Metric Optimised For\\$\rightarrow$ Labelled Metric}} & $l-R^2$ & $q-R^2$ & $W$ & $MAE$ \\
        \cline{2-5}
        & \multicolumn{4}{c}{Default Hyperparameters ($FCR:\: 1.751$)} \\
        \midrule
        \midrule
        $\mathcal{M} \rightarrow F1$ & 0.54 & 0.59 & 0.15 & -0.01 \\
        $\mathcal{M} \rightarrow NMI$ & 0.52 & 0.62 & 0.13 & -0.05 \\
        $\mathcal{C} \rightarrow F1$ & 0.26 & 0.28 & 0.19 & -0.01 \\
        $\mathcal{C} \rightarrow NMI$ & 0.29 & 0.32 & 0.16 & -0.04 \\
        \midrule
         & \multicolumn{4}{c}{Hyperparameter Optimisation ($FCR:\: 1.249$)} \\
        \midrule
        \midrule
        $\mathcal{M} \rightarrow F1$ & 0.49 & 0.58 & 0.22 & -0.01 \\
        $\mathcal{M} \rightarrow NMI$ & 0.5 & 0.62 & 0.21 & -0.03 \\
        $\mathcal{C} \rightarrow F1$ & 0.01 & 0.13 & 0.26 & -0.15 \\
        $\mathcal{C} \rightarrow NMI$ & 0.03 & 0.06 & 0.28 & -0.13 \\
        \bottomrule
    \end{tabular}
    \caption{We compare the performance of algorithms trained without any labelled comparison for model selection under the default hyperparameters as well as for hyperparameter optimisation. The correlation between the unsupervised metrics optimised for and the measures of performance on the labels is given with both a linear and quadratic fit, where the $R^2$ coefficient of determination is given to quantify the strength predictability on a scale of $0$ to $1$. We also report the difference in performance on the ground-truth metric when the ground-truth metric was used for the optimisation. The algorithm's ranking sensitivity to randomness is also given for each metric on a scale of $0$ to $1$. To compare the absolute performance of the two hyperparameter setups, the framework comparison rank (FCR) is used \citep{ugle2023leeney} to show that the HPO returned better performance than the default set of parameters.}
    \label{tab:def_hpo_overall_res}
\end{table}

\renewcommand{\arraystretch}{1.2}
\setlength{\tabcolsep}{10pt}
\begin{table}[h!tb]
  \centering
  \begin{minipage}{\linewidth}
    \centering
    \small
    \captionsetup{font=small} 
    \begin{tabular}{c|cccc}
        \toprule
        q-$R^2$ Default/(HPO) & $\mathcal{M} \rightarrow F1$ & $\mathcal{M} \rightarrow NMI$ & $\mathcal{C} \rightarrow F1$ & $\mathcal{C} \rightarrow NMI$ \\
        \midrule
        \multicolumn{5}{c}{\textbf{Algorithm Correlation's}} \\
        \midrule
        \midrule
        DGI & 0.76(\textbf{0.77}) & 0.85(\textbf{0.86}) & 0.23(0.23) & \textbf{0.29}(0.12) \\
        DAEGC & \textbf{0.78}(0.67) & \textbf{0.81}(0.78) & \textbf{0.83}(0.33) & \textbf{0.73}(0.3) \\
        DMON &  0.65(\textbf{0.68}) & \textbf{0.84}(0.81) & 0.21(\textbf{0.49}) & 0.30(\textbf{0.43})  \\
        GRACE & \textbf{0.66}(0.65) & 0.59(\textbf{0.72}) & \textbf{0.18}(0.03) & \textbf{0.17}(0.06) \\
        SUBLIME & 0.61(\textbf{0.75}) & 0.50(\textbf{0.7}) & \textbf{0.53}(0.18) & \textbf{0.38}(0.21) \\
        BGRL & 0.57(\textbf{0.62}) & 0.69(0.69) & \textbf{0.13}(0.03) & \textbf{0.29}(0.15)  \\
        VGAER & \textbf{0.90}(0.5) & \textbf{0.69}(0.5) &\textbf{0.76}(0.37) & 0.25(\textbf{0.33})  \\
        \midrule
        \multicolumn{5}{c}{\textbf{Dataset Correlation's}} \\
        \midrule
        \midrule
        CiteSeer & \textbf{0.50}(0.37) & \textbf{0.62}(0.26) & 0.09(\textbf{0.12}) & \textbf{0.30}(0.12) \\
        Cora & 0.18(\textbf{0.51}) &  \textbf{0.59}(0.41) & 0.05(\textbf{0.11}) & \textbf{0.41}(0.10 ) \\
        Texas & 0.00(\textbf{0.03}) & 0.06(0.06) & 0.21(\textbf{0.70}) & 0.12(\textbf{0.56}) \\
        DBLP & 0.33(\textbf{0.52}) &  0.36(\textbf{0.40}) & 0.12(\textbf{0.39}) & \textbf{0.28}(0.23) \\
        Wisc & 0.02(\textbf{0.03}) & \textbf{0.02}(0.01) & 0.12(\textbf{0.69}) & 0.03(\textbf{0.45}) \\
        Cornell & 0.01(\textbf{0.25}) & 0.07(\textbf{0.36}) & 0.00(\textbf{0.49}) & 0.02(\textbf{0.42}) \\
        \bottomrule
    \end{tabular}
    \caption{The $R^2$ Coefficient of Determination of the quadratic regression given between the GNNs optimised under Modularity or Conductance compared with the metrics NMI and F1. Here the regression is performed on each dataset or algorithm individually to quantify the predictability of each subset of the tests. Highlighted in bold is the stronger correlation between default and optimised hyperparameter's for each algorithm or dataset.}
    \label{tab:def_hpo_dataset_algos}
  \end{minipage}
\end{table}

In Table \ref{tab:def_hpo_overall_res}, we can see that there is a correlation between the models selected with modularity compared with the ground-truth performance, however this is not present for conductance. The low value for $MAE$ indicates that the models found using modularity versus directly selecting for $F1$ and $NMI$ are insignificantly worse; under the $\mathcal{M} \rightarrow F1$ prediction, the $F1$ score is only $0.01$ worse averaged over all benchmark tests. The strongest correlation of prediction is from $\mathcal{M} \rightarrow NMI$ under the quadratic model and interestingly, these are also the most least sensitive to randomness. Figure \ref{fig:hpo_def_corr} agrees with the assessment that conductance is not as good a predictor of labeled performance. This is because as conductance decreases, the labelled performance spreads out, telling us that this could lead to better performance but there is no confidence in this predictor. In contrast, both models of prediction for modularity show that as modularity increases the same is seen for both the ground-truth indicators. 

We also investigate whether these correlations are still present when using unsupervised metrics for hyperparameter optimisation as well as model selection. However, in general, there is less correlation with this additional challenge. When comparing the absolute performance with the $FCR$ we do find that the hyperparameter optimisation still leads to better performance than simply using the default values. Additionally, there is not that much difference between a ground-truth supervised HPO compared to the unsupervised one that we propose in this investigation as for modularity there is only a drop of $0.01$ for $F1$ and $0.03$ for $NMI$ over all datasets and algorithms. Conductance is completely uncorrelated with performance on the labels for each dataset when carrying out the HPO. Modularity is still a good indicator of performance for both $NMI$ and $F1$, however the $W$ randomness coefficient has increased which means that the ranking of algorithm's is less certain and therefore less trustworthy.

Given that there is a meta trend of correlation between unsupervised GNNs and ground-truth performance, we can also look at the individual correlations for each algorithm and dataset. The quadratic model is used in Table \ref{tab:def_hpo_overall_res} as the trends in Figures \ref{fig:hpo_def_corr} all exhibit more non-linear trends as given by the higher $R^2$ correlation. From Table \ref{tab:def_hpo_overall_res}, the trend of Modularity being more correlated than Conductance for all of the results still holds true in most scenarios. The boldest result is simply the strongest correlation for each algorithm or dataset consideration under the HPO or default HPs. It can be seen that by comparing this to the trend for the whole framework, each algorithm is more predictable by itself than the trend of the entire suit of models. VGAER is the most correlated on a single metric pair closely followed by DGI, DMON and DAEGC. This suggests that some algorithm architectures are better for unsupervised optimisation with modularity. 

The individual datasets correlations are always lower than the whole frameworks trend. We can see that the individual dataset performance correlation is increased by the hyperparameter optimisation under the $\mathcal{M} \rightarrow F1$ prediction scenario for all datasets bar the CiteSeer dataset. It is important to note that some datasets have no correlation between unsupervised and labelled performance even using modularity. We also look at the absolute performance comparison between the default and hyperparameter optimisation GNNs to make sure than as well as a present correlation, that they are also learning. Comparing Table \ref{tab:def_hpo_dataset_algos} to Figure \ref{fig:def_hpo_abs}, we can see that the datasets for which there is a higher correlation (Cora, CiteSeer, DBLP) also train better. Those that have a weak correlation (Texas, Wisc, Cornell) also perform badly, which could be because these datasets are small. Interestingly, if we look at the dataset information in Table \ref{tab:stats}, we can see that the datasets with poor performance have more feature space information, whereas the good dataset performance is correlated with more adjacency space information. 


\subsection{Reduced Training Dataset Size (RQ4)}

We have also investigated the extent to which the predictability of unsupervised algorithms is affected by the training dataset size. This helps us to measure how much data is needed to determine a good model (RQ4). We use the default hyperparameters and train on a subset of the original dataset by reducing the number of edges for training to compare performance of the unsupervised metrics on the reduced dataset to the labelled performance on the original dataset. 

From Table \ref{tab:percent} it can be seen that there is still a correlation when there is less data available and again the quadratic model has a higher correlation. Also, in consistency with the full data availability, modularity is better than conductance for predicting supervised metric performance from the reduced set. There is however a decrease in correlation when reducing to $66\%$ but from there almost none in the drop to $33\%$. This indicates that dataset size does impact the correlation in performance and may suggest that using more unlabelled data is not always a good idea. In opposition to this stance, the $MAE$ comparing supervised performance under a supervised model selection indicates that there is not a significant drop in performance. This suggests that the drop in predictability is because of the reduced training data rather than the unsupervised optimisation. 

In Table \ref{tab:dmon_large} we test the predictability of unsupervised metrics on larger datasets to investigate if the conclusions scale. The only algorithm tested in this paradigm is DMON due to constraints with the runtime and this algorithm was the fastest implemented. We optimise the hyperparameters with the same constraints as the previous experiment. It can be seen that the absolute performance indicates that the algorithm did learn on both datasets when modularity is used for model selection and HPO. However, Photos was significantly predictable and also has better performance than the Computers dataset. From this, we cannot confidently conclude that this scales or not as there is not enough evidence for either. This is also because there is not an insignificant drop in both $NMI$ and $F1$ when using $\mathcal{M}$ in the optimisation process.

\renewcommand{\arraystretch}{1.2}
\begin{table}[htb!]
\centering
\captionsetup{font=small} 
\small
    \begin{tabular}{c|cccc}
        \toprule
        \multirow{2}{*}{\Longunderstack{Metric Optimised For\\$\rightarrow$ Labelled Metric}} & $l-R^2$ & $q-R^2$ & $W$ & $MAE$ \\
        \cline{2-5}
        & \multicolumn{4}{c}{$66\%$ Training Data} \\
        \midrule
        \midrule
        $\mathcal{M} \rightarrow F1$ & 0.43 & 0.48 & 0.16 &  -0.03\\
        $\mathcal{M} \rightarrow NMI$ & 0.44 & 0.52 & 0.13 & -0.03\\
        $\mathcal{C} \rightarrow F1$ & 0.14 & 0.16 & 0.17 & -0.05 \\
        $\mathcal{M} \rightarrow NMI$ & 0.17 & 0.17 & 0.15 & -0.05 \\
        \midrule
        & \multicolumn{4}{c}{$33\%$ Training Data} \\
        \midrule
        \midrule
        $\mathcal{M} \rightarrow F1$ & 0.39 & 0.42 & 0.15 & -0.05 \\
        $\mathcal{M} \rightarrow NMI$ & 0.41 & 0.46 & 0.13 & -0.04\\
        $\mathcal{C} \rightarrow F1$ & 0.15 & 0.16 & 0.15 & -0.06\\
        $\mathcal{C} \rightarrow NMI$ & 0.20 & 0.20 & 0.15 & -0.05\\
        \bottomrule
    \end{tabular}
    \caption{The $R^2$ Coefficient of Determination Linear and Quadratic given between GNNs optimised on $33\% \mathbin{/} 66\%$ of the total data available using Modularity or Conductance, compared with the performance on NMI and F1 on all of the data. We also detail the absolute difference in performance from the same setup, except where NMI and F1 are used for model selection. }
    \label{tab:percent}
\end{table}
\newpage

\begin{table}[h!btp]
\centering
\small
\captionsetup{font=small} 
    \begin{tabular}{c|cc}
        \toprule
        \Longunderstack{Large Datasets\\(q-$R^2$/abs/$MAE$)}& Computers & Photo \\
        \midrule
        \midrule
        $\mathcal{M} \rightarrow F1$ & 0.22/0.35/-0.08 & 0.98/0.57/0.16 \\
        $\mathcal{M} \rightarrow NMI$ & 0.33/0.49/-0.02 & 0.98/0.56/-0.09 \\
        $\mathcal{C} \rightarrow F1$ & 0.84/0.09/-0.34 & 0.91/0.06/-0.66 \\
        $\mathcal{C} \rightarrow NMI$ & 0.75/0.06/-0.45 & 0.99/0.02/-0.63 \\
        \bottomrule
    \end{tabular}
  \caption{In this experiment, the DMON algorithm is trained on larger datasets. The hyperparameters are optimised using Conductance and Modularity as well as being used for model selection. The first metric is the $R^2$ from the quadratic is fitted to assess ``ground-truth" predictability, the absolute ground-truth metric performance is next, then the  absolute difference in performance where NMI and F1 are used for model selection and HPO. }
  \label{tab:dmon_large}
\end{table}

\begin{figure*}[hp!]
    \centering
    \small
    \captionsetup{font=small} 
    \begin{minipage}[b]{\textwidth}
        \centering
        \captionsetup{font=small} 
        \includegraphics[width=\textwidth,height=0.38\textheight]{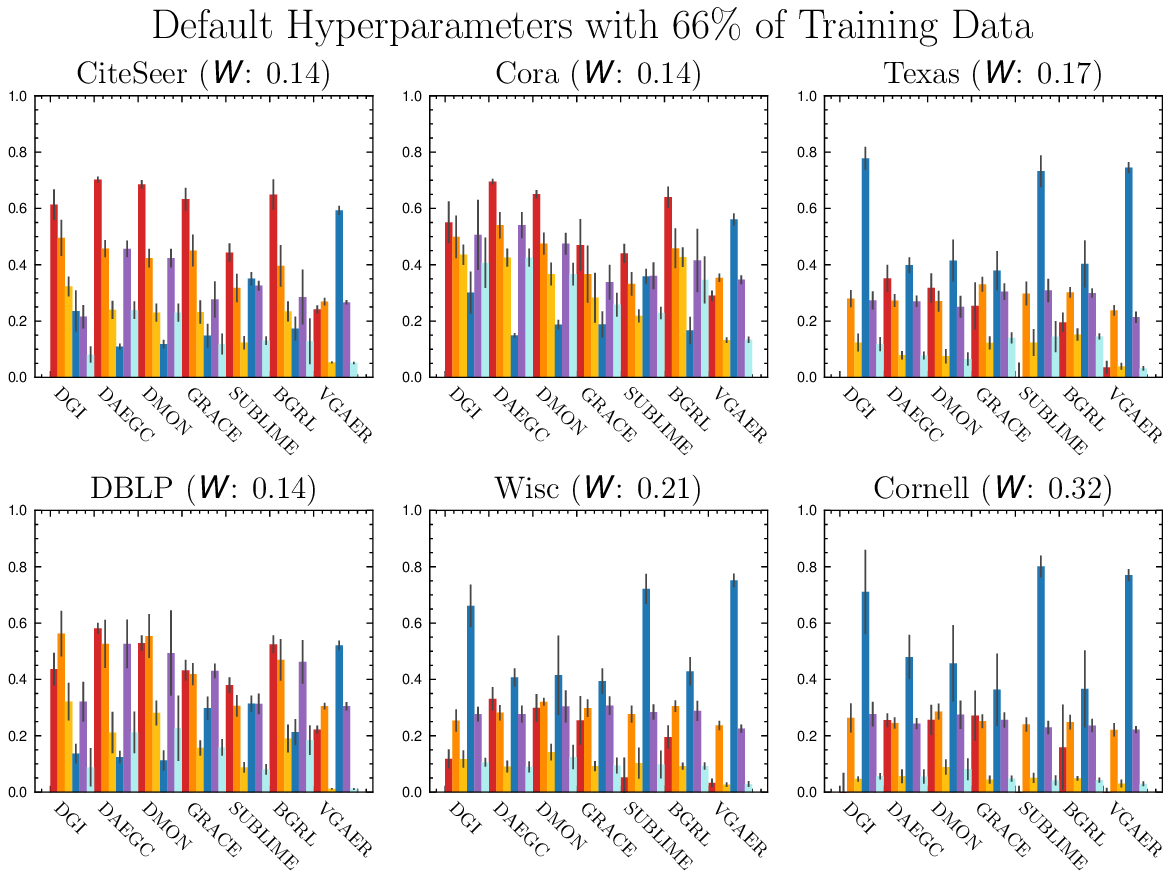}
    \end{minipage}
    \vspace{0.8cm}
    \begin{minipage}[b]{\textwidth}
        \centering
        \captionsetup{font=small} 
        \includegraphics[width=\textwidth,height=0.38\textheight]{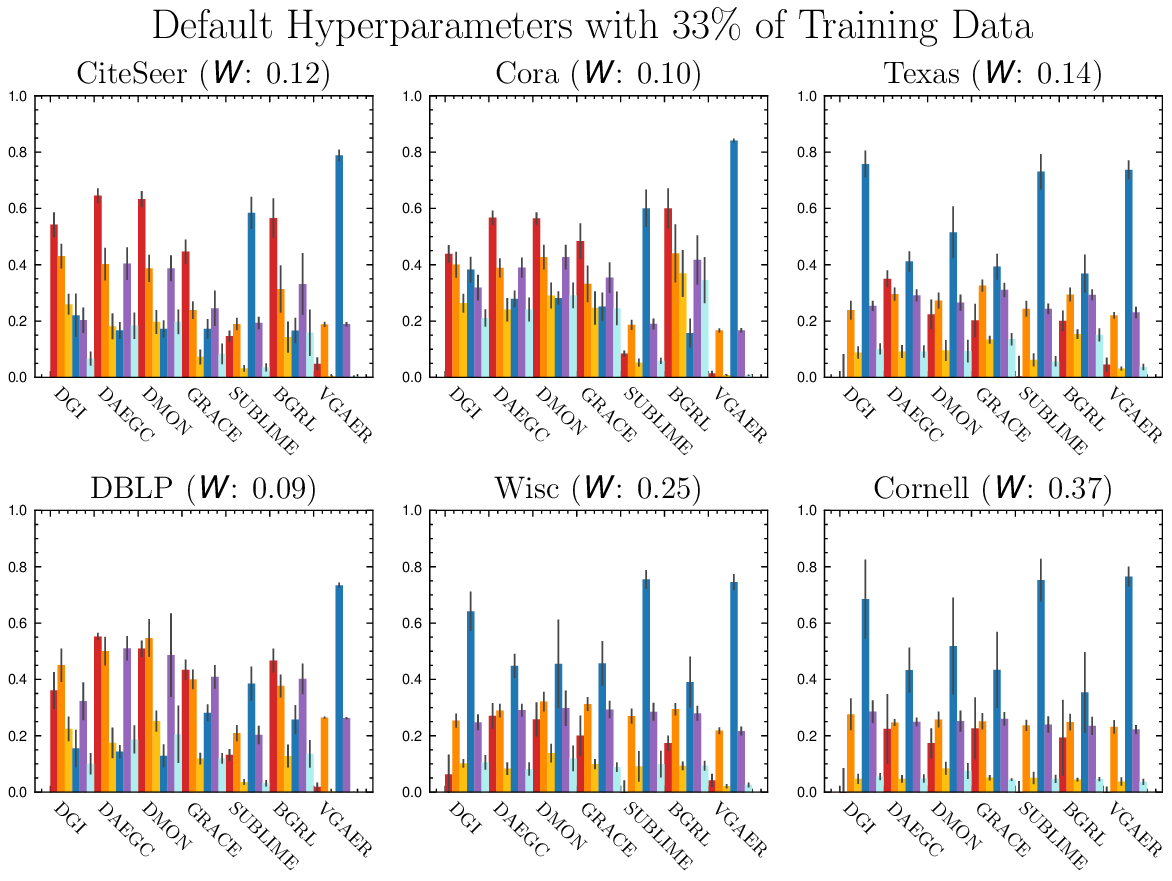}
        \includegraphics[height=0.07\textheight]{handles_abs.eps}
        \caption{ Each algorithm is trained on 66\% and 33\% and of the edges available in the original dataset on the ground-truth and given is the absolute performance of the ground-truth F1 and NMI metrics whilst using the default hyperparameters with model selection based on Modularity and Conductance.}
        \label{fig:def_hpo_abs_percent}
    \end{minipage}
\end{figure*}


\newpage
\subsection{Synthetic Datasets (RQ5)}

\begin{figure}[!bhp]
\captionsetup{font=small} 
    \begin{center}
        \includegraphics[width=\linewidth,height=0.75\textheight]{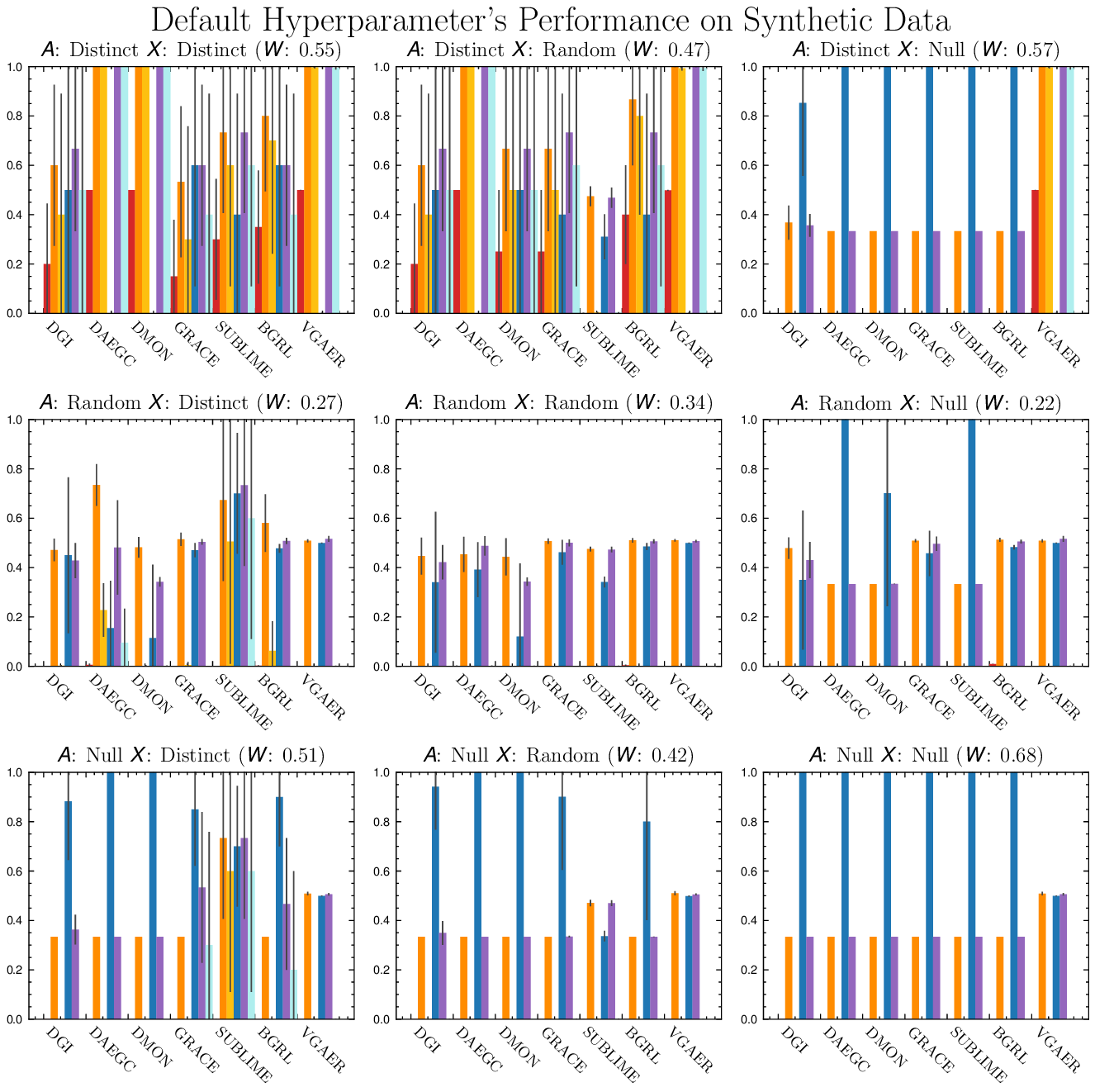}
        \includegraphics[height=0.07\textheight]{handles_abs.eps}
        \caption{The values of NMI and F1 of each GNN clustering algorithm when optimised under modularity and conductance on different synthetic clustering scenarios. Each algorithm is optimised on multiple seeds and the average performance across all is given. There are 1000 nodes and 500 features in all scenarios. The distinct scenario is complete separation between clusters; random means that it is completely random between clusters; the null means that there is explicitly no information distinction between the clusters.}
        \label{fig:synth}
    \end{center}
\end{figure}

In Figure \ref{fig:synth}, it can be seen that not all algorithms can perfectly discern the easy ground-truth clustering in an attributed graph by using modularity or conductance. This is subject to randomness as it can be seen that the performance is significantly different depending on the random seed. Only DAEGC, DMON and VGAER can perfectly discern the clustering partition every time. BGRL and SUBLIME do so subject to the random seed, whereas DGI and GRACE do not cluster the data perfectly under any seed. It is surprising that every algorithm does not cluster the data in this scenario as the clusters are completely distinct in both information spaces. This is not due to the unsupervised optimisation as from Table \ref{tab:synth_diff} we can see that there is little drop in performance compared to using the supervised metrics for optimisation. However, this may be because the adjacency matrix is completely disconnected so information learnt can't be shared between the two clusters as the computation graph will also be detached \citep{alon2020bottleneck}. Although the algorithms that aren't reliable are those that employ a corruption function for training, which may also be the reason for unreliable performance. 

\setlength{\tabcolsep}{6.8pt}
\begin{table}[!htb]
\centering
\small
\captionsetup{font=small} 
    \begin{tabular}{c|cccc}
        \toprule
        Synthetic Datasets & Mod $\rightarrow$ F1 & Mod $\rightarrow$ NMI & Con $\rightarrow$ F1 & Con $\rightarrow$ NMI \\
        \midrule
        \midrule
        \textbf{A}: Distinct \textbf{X}: Distinct & 0.04 & -0.01 & 0.03 & -0.03 \\
        \textbf{A}: Distinct \textbf{X}: Random & 0.01 & -0.04 & 0.01 & -0.04 \\
        \textbf{A}: Distinct \textbf{X}: Null & 0.0 & 0.0 & 0.0 & 0.0 \\
        
        \textbf{A}: Random \textbf{X}: Distinct & -0.07 & -0.15 & -0.13 & -0.17 \\
        \textbf{A}: Random \textbf{X}: Random & -0.01 & 0.0 & -0.03 & 0.0 \\
        \textbf{A}: Random \textbf{X}: Null & -0.01 & 0.0 & -0.01 & 0.0 \\
        
        \textbf{A}: Null \textbf{X}: Distinct & -0.04 & -0.11 & 0.01 & -0.04 \\
        \textbf{A}: Null \textbf{X}: Random & -0.02 & -0.03 & -0.02 & -0.03 \\
        \textbf{A}: Null \textbf{X}: Null & 0.0 & 0.0 & 0.0 & 0.0 \\ 
        \bottomrule
    \end{tabular}
  \caption{The absolute difference in performance from using modularity or conductance for model selection compared to using NMI or F1 for the synthetic datasets experiment is shown.}
  \label{tab:synth_diff}
\end{table}

When the feature space is random but the adjacency space is still disjoint only SUBLIME and DGI fail to cluster under any random seed. Every other algorithm does so at least once, which is surprising given that GRACE failed to cluster perfect in the easier scenario of both spaces being distinct. Comparing to the synthetic scenarios where the clusters are present in the feature space, more algorithms fail to cluster based on unsupervised information. This suggests that node clustering performed with GNNs is biased towards clusters that appear in the adjacency space due to the use of a graph partitioning quality metric so consequently are under-utilising the feature space.

In the null feature space clustering partition with disjoint adjacency space, the VGAER algorithm can perfectly cluster, which means that it is not properly balancing the uncluster-able signal from the feature space. Another interesting observation of the VGAER is that it never performs worse than random in any scenario, whereas in the scenario of no cluster-able information in either space, all other algorithms perform worse than random. This might be because this is the only GNN variant that assumes a distribution in the latent space.

\section{Discussion}

The findings indicate a correlation between the performance of unsupervised metrics used in optimising GNNs and the ground-truth metrics, showing that modularity can be used in model selection for GNN clustering algorithms. Therefore the answer to RQ1 is yes. This correlation is important as unsupervised learning without any prior information can be used where labeled data is scarce or unavailable. Unsupervised learning can be valuable technique for pre-training models \citep{erhan2010does} though not explicitly showcased in this work, our work implies the utility of pre-training despite a direct demonstration, which is particularly useful in scenarios necessitating initial data collection in the absence of available data. The extent of this can be quantified by further investigations. 

The uncertainty in the results is indicated by the high $W$ randomness coefficient, which quantifies how much the ranking of different algorithms changes under the presence of randomness. This gives an estimation for the amount of randomness affecting the results; if the rankings change a lot between different seeds, then the performance of each algorithm will also be subject to a lot of change between seeds. Modularity has a higher correlation with the ground-truth performance in contrast to conductance. Using modularity as the predictor had a lower $W$ randomness coefficient than conductance which means these results are more reliable. A good modularity score does predict ground-truth well but this is subject to a high amount of variation. Therefore, whilst the ground-truth be predicted, this is an estimate rather than an absolute statement of performance. 

In regards to RQ2, we find that when using modularity, an unsupervised optimisation doesn't lead to a significant drop in performance relative to a labelled comparison. However, conductance has not been shown to predict the ground-truth performance which means that not all unsupervised metrics are sufficient for predictability. Good unsupervised learning is important in of itself but it is also useful to use well trained unsupervised models for effective transfer learning \citep{sener2016learning}. There are more unsupervised metrics that can be compared than those tested, which future investigations can explore. 

To answer RQ3, on the optimisation of hyperparameters using unsupervised metrics, we show that modularity can be used to train unsupervised GNN models to cluster in absence of known good hyperparameters. It is also shown that the hyperparameters found increase performance relative to the default values. This is important as previously, in practical applications for new datasets without label access, hyperparameters could not be tuned to increase performance. Without knowledge of good hyperparameters, this work shows that it is possible to train unsupervised GNNs.

Concerning RQ4, we find that reducing the dataset size for training compared to the evaluation size changes the strength of the correlation in performance. There is still a correlation when using modularity and the $MAE$ wasn't significantly different from a supervised model selection so the drop in predictability is likely because of the reduced training set. When we increased the size of the dataset we did not find any consistent findings of predictability from the unsupervised optimisation. We also found that each individual dataset correlations are lower than the average of all tests of the framework whereas the algorithm predictability is much higher. This means that it is easier to predict how well an algorithm will do on any given dataset than how easy it is to extract information for a particular dataset. We find that datasets with a higher ratio of graph information are more predictable and algorithms performed better on these. This is likely because modularity is a graph partitioning quality metric and use of this metric is biased towards clusters in the graph-structure. We can interpret this as meaning it is more worthwhile to collect graph data or labels for feature heavy datasets.

The last set of experiments pertain to RQ5, which is why does training GNNs to cluster with modularity work? The synthetic experiments show that GNNs are under utilising the duality of the information spaces when trained with an unsupervised pipeline. Only three GNN variants can reliable cluster when both information spaces are disjoint (DAEGC, DMON, VGAER). It is surprising that others are subject to randomness and are therefore less reliable algorithms. Some, such as DGI, GRACE, SUBLIME and BGRL fail to consistently cluster the completely disjoint clustering partition using unsupervised optimisation of the models but this may be because they are dependent on corruption functions. Most algorithms in the adjacency distinct and features random partition can cluster but not consistently across all random seeds. It is not surprising that this is not reliable as the random information from the feature space adds noise to the clustering signal. However, in the reverse scenario of distinct feature information and random connections, no algorithm can reliably cluster. This was the scenario with the greatest drop in performance compared to a supervised evaluation. The reduced performance in this scenario is likely because the graph partitioning metrics only measure the clustering quality with respect to the adjacency space, whereas the predominant clustering signal here comes from the features. This means that when using modularity in lieu of ground-truth, the adjacency space needs to reflect the potential ground-truth. Future work could investigate using unsupervised clustering quality partitioning metrics in the feature space to mitigate this. 

\section{Conclusion}

In this work, it is found that modularity can be used to train unsupervised GNN models and perform model selection, with performance serving as a proxy for evaluating against the ground-truth. It is shown that modularity can be used to optimise hyperparameters for better clustering performance than the default hyperparameters, without a significant drop relative to a supervised investigation. We find that the lower limit of data needed to optimise GNNs for clustering in performance beyond the training data depends on the graph structured information. Using synthetic data partitions we find that the limiting factor of unsupervised GNN optimisation is that they are biased towards connectivity space cluster partitions as we use a modularity objective.

\vskip 0.2in
\bibliographystyle{plainnat}
\bibliography{references}

\end{document}